\documentclass[letterpaper]{article}
\usepackage[preprint]{aaai2027}
\usepackage[hyphens]{url}
\usepackage{graphicx}
\usepackage{natbib}
\usepackage{booktabs}
\usepackage{amsmath}
\usepackage{amssymb}

\makeatletter
\newcommand{\captionof}[1]{\def\@captype{#1}\caption}
\makeatother
\newcommand{\sys}{SearchWiki}
\newcommand{\ours}{WikiResearcher-9B}
\newcommand{\tablesize}{\small}
\newcommand{\tablepad}{\setlength{\tabcolsep}{4pt}}

\title{\sys: Learning to Build and Navigate Knowledge Wikis for Active Information Seeking}

\author {
    Guransh Singh\textsuperscript{\rm 1}\thanks{Work done as an intern at IBM},
    Vishwajeet Kumar\textsuperscript{\rm 2},
    Arkadeep Acharya\textsuperscript{\rm 2},
    Adnan Qidwai\textsuperscript{\rm 2},
    Jaydeep Sen\textsuperscript{\rm 2},
    Sachindra Joshi\textsuperscript{\rm 2}
}
\affiliations {
    \textsuperscript{\rm 1}Birla Institute of Technology, Mesra\\
    \textsuperscript{\rm 2}IBM\\
    guransh766@gmail.com,  vishk024@in.ibm.com, acharyarka17@ibm.com, Adnan.Qidwai1@ibm.com, jaydesen@in.ibm.com, jsachind@in.ibm.com
}
\usepackage{color}

\newcommand{\eat}[1]{} 

\eat{}

\begin{document}

\maketitle

\begin{abstract}
Flat retrieval-augmented generation treats a corpus as a bag of chunks, discarding document hierarchy and cross document structure.
We introduce \textbf{\sys}, a harness framework that synthesizes a corpus into a hierarchical, typed, navigable wiki and trains an agent, \textbf{\ours}, to retrieve information through multi-turn tool use.
The wiki organizes knowledge into three layers - document overviews, cross-document topic pages, and page-level source records; enabling progressive refinement of retrieval when initial lookup misses.
We optimize the agent's navigation policy with on-policy reinforcement learning with a multi-component reward function balancing answer correctness, retrieval quality and trajectory efficiency.
Evaluation on ViDoRe-V3 (8 domains), FinanceBench, and memory benchmarks (LoCoMo, LongMemEval, PersonaMem-v2) shows that \textbf{\ours} which is our RL-tuned Qwen 9B model, significantly outperforms same-size untrained baselines and exceeds or matches larger external models. \textbf{\sys} paired with \textbf{\ours} demonstrates that learned navigation over structured corpora is a superior alternative to flat retrieval.

\end{abstract}

\section{Introduction}
Retrieval augmented generation (RAG) has emerged as a dominant paradigm for grounding large language models on external knowledge~\cite{lewis2020rag,gao2023ragsurvey}, mitigating hallucination~\cite{shuster2021retrieval} and enabling access to information beyond model's parametric memory or training cutoff. In its canonical form, RAG operates by embedding a corpus into a vector space, retrieving the top-k chunks most similar to a query~\cite{karpukhin2020dpr}, and conditioning generation on this retrieved context - an approach that has proven remarkably effective for open domain question answering(QA), enterprise search, and knowledge intensive/grounded NLP tasks more broadly. However, the "flat" formulation of RAG treats corpus as an undifferentiated bag of chunks, indexed and retrieved independently of one another. In doing so, it discards the very structure that makes the corpus navigable: the hierarchy of chapter, sections, sub-sections and the cross references and the citations linking the documents and the implicit organization that a human reader relies on to build context incrementally~\cite{liu2021densehierarchical,sarthi2024raptor,edge2024graphrag,wang2024hipporag}. Critically, flat RAG offers no recovery mechanism when initial retrieval fails - if the top-k chunks do not contain the answer, there is no mechanism for the system to reformulate its search, backtrack, or explore related regions of the corpus, since retrieval is a single-shot operation than an iterative process. The agent, in effect cannot navigate; it can only query and hope. As the corpora grow larger and more complex spanning technical documentations, legal codices, financial reports or scientific literature with deep hierarchical dependencies, answering non-trivial questions increasingly requires multi-step, structured reasoning that flat RAG/retrieval architectures are fundamentally unequipped to provide~\cite{trivedi2023ircot,jiang2023flare,asai2024selfrag,jeong2024adaptiverag,yao2023react}.
 \par Consider how a human researcher approaches an unfamiliar body of knowledge: not by scanning every paragraph in isolation, but by navigating hierarchically starting from an overview of table of contents, drilling down into relevant chapters, sections, subsections, concepts and following cross references until sufficient detail is gathered to answer question at hand. This process is inherently structured and recoverable: if a chosen path proves unproductive, the reader backtracks to a higher level and tries another branch rather than abandoning the search 
entirely~\cite{bates1989berrypicking,pirolli1999informationforaging}. Flat RAG systems, by contrast have no notion of "levels" to retreat to or "branches" to explore, every chunk is equally proximate to every query, there is no scaffold on which the agent might orient itself. This raises a natural question: why doesn't RAG simply adopt the same navigational strategy as humans? Our insight is that this capability can be manufactured rather than assumed. Specifically, we propose to synthesize a wiki-like hierarchical structure directly from the unstructured corpus automatically by inducing overview pages, concept level summaries, cross document links~\cite{ming2026llmwiki,dai2026navirag,du2026arag,sun2026corpus2skill,xu2026napmem} and train an agent via reinforcement learning (RL), to navigate the synthesized structure much as a human would traverse a wiki. We argue that information seeking over corpora should move from \textbf{passive retrieval} to \textbf{active corpus navigation}~\cite{irag2025,grepseek2026}: the agent should choose what to inspect, in what order, and at what granularity, using intermediate observations to decide the next step.
\par We instantiate this insight into two-stage stage framework called \textbf{\sys}, First, \textit{synthesis:} an LLM processes the raw corpus and builds a typed, three-layer wiki:
\begin{itemize}
    \item \textbf{Document overviews (\textit{documents/}):} per-document summaries and optional section subfolders that act as entry points and proxies to direct sources.
    \item \textbf{Topics (\textit{topics/}):} cross-document concept pages with markdown links to related documents and topics, to bridge the cross-document and cross-section information gap.
    \item \textbf{Sources (\textit{sources/}):} page-level JSON records (OCR/layout text and related metadata) for fine-grained verification and cited answering.
\end{itemize}
This layout supports \emph{progressive disclosure}: the agent can start from summaries or topics and descend to specific pages only when needed.
The same tool suite (\textit{search\_wiki}, \textit{read\_file}, \textit{read\_page}, \textit{grep\_search}, and related filesystem tools) operates over any wiki compiled into this convention.
Second, \textit{learning:} we train \textbf{\ours}, an agent via RL to navigate this wiki, deciding when to ascend into details and when to backtrack, conditioned on the query. This works because it replaces passive, single-shot retrieval with a learned policy operating over explicit structure: the wiki provides a navigable scaffold, while RL training gives the agent the judgment to use it efficiently and recover from dead ends. Two design choices matter most: the three-layer hierarchy balances global orientation against local precision, and a multi-component reward jointly optimizes for answer correctness, navigation efficiency, and effective backtracking.

We train the navigation policy with on-policy RL using Decoupled Clip and Dynamic Sampling Policy Optimization~\cite{yu2025dapo} (DAPO), building on group-relative advantages introduced by GRPO~\cite{shao2024grpo}.
Rollouts are multi-turn tool trajectories over live wiki environments.
\textbf{Retrieved Token Masking (RTM)}~\cite{jin2025searchr1} restricts policy gradients to model-generated tokens (planning, tool calls, and answers), not tool observations.

Our contributions are threefold:
\begin{enumerate}
    \item A domain-agnostic wiki compilation and navigation interface for any knowledge source, with progressive disclosure over document, topic, and source layers.
    \item An online RL training recipe (DAPO, RTM, retrieval-aligned dense rewards) for tuning multi-turn agents such as \ours{} to learn an effective search strategy.
    \item A unified evaluation across all eight publicly released ViDoRe~v3 splits\cite{vidorev3}, comparing \ours{} with untrained baselines and larger external models under a common harness, together with out-of-domain evaluation on FinanceBench, LoCoMo, LongMemEval, and PersonaMem-v2~\cite{islam2023financebench,maharana2024locomo,wu2024longmemeval,jiang2025personamemv2}.
\end{enumerate}

The rest of the paper is organized as follows.
Section~\ref{sec:related} discusses related approaches.
Section~\ref{sec:wikiresearcher} presents the \sys{} wiki representation, ingestion process, tool interface, and DAPO training recipe.
Section~\ref{sec:experiments} describes the experimental setup.
Section~\ref{sec:result_analysis} reports ViDoRe~v3 results, ablations, generalization holdouts, and analyses when navigation helps.
Section~\ref{sec:conclusion} concludes and summarizes the manuscript.

\section{Related Work}
\label{sec:related}

\paragraph{Retrieval-augmented generation.}
RAG systems retrieve top-$k$ passages and feed them to a generator.
ViDoRe~\cite{faysse2025vidore,vidorev3} stresses retrieval over visually complex, multi-page documents.
\sys{} differs by treating retrieval as an interactive search-and-read loop over a pre-compiled wiki, rather than a single flat context pack.

\paragraph{Agentic search and reinforcement learning.}
Search-R1~\cite{jin2025searchr1} and DeepRetrieval~\cite{jiang2025deepretrieval} train agents to issue search queries.
GrepSeek~\cite{grepseek2026} trains agents to operate on raw text with Unix-style tools.
Interact-RAG~\cite{irag2025} exposes fine-grained corpus operations as interactive primitives.
KARL~\cite{karl2026}, DeepRAG~\cite{deeprag2026} and GraphRAG-R1~\cite{yu2026graphragr1} further study RL for retrieval decisions.
TMAX~\cite{tmax2026} shows that simple terminal rewards can suffice in constrained agent settings.
\sys{} uses a different \emph{search space}: a typed, structured wiki (document summaries, cross-document topics, and page-level sources) that the agent traverses with domain-agnostic tools, rather than a flat search API or an unstructured text dump.

\paragraph{Structured memory versus corpus wikis.}
Memory systems for personal assistants organize user histories as records, graphs, or hierarchies~\cite{zhong2023memorybank,packer2024memgpt,chhikara2025mem0}.
AgeMem~\cite{yu2026agemem} and NapMem~\cite{xu2026napmem} train agents to navigate multi-layer \emph{user memory}, with NapMem specializing its RL policy on memory benchmarks such as LoCoMo, LongMemEval, and PersonaMem-v2. Similarly, PageIndex~\cite{zhang2025pageindex}, proposes vectorless RAG by constructing a hierarchical tree index from long documents;
Inspired by LLM-wiki~\cite{karpathy2026llmwiki}, \sys{} aims to convert a given \emph{document corpora} to a domain-agnostic wiki format~\cite{ming2026llmwiki} and provides a harness with an effective tool set to traverse the wiki efficiently.
Given the same wiki representation, a \sys{} harness using both an untrained large as well as our ViDoRe-trained \ours{} model outperform NapMem's reported memory-specialized numbers on key metrics on LoCoMo (and the harness also leads on PersonaMem-v2; Table~\ref{tab:memory}), suggesting that structured navigation transfers beyond the training domain.

\paragraph{RL for tool-using language models.}
GRPO~\cite{shao2024grpo} estimates advantages from group-normalized rewards without a critic.
DAPO~\cite{yu2025dapo} uses asymmetric clipping and dynamic sampling for long-trajectory stability.
RTM-style masking~\cite{jin2025searchr1} avoids updating the policy on environment tokens.
We combine these ingredients with a \sys-specific dense reward stack.

\section{\sys}
\label{sec:wikiresearcher}


\subsection{Overview}
\begin{figure*}[t]
\centering
\includegraphics[width=\textwidth]{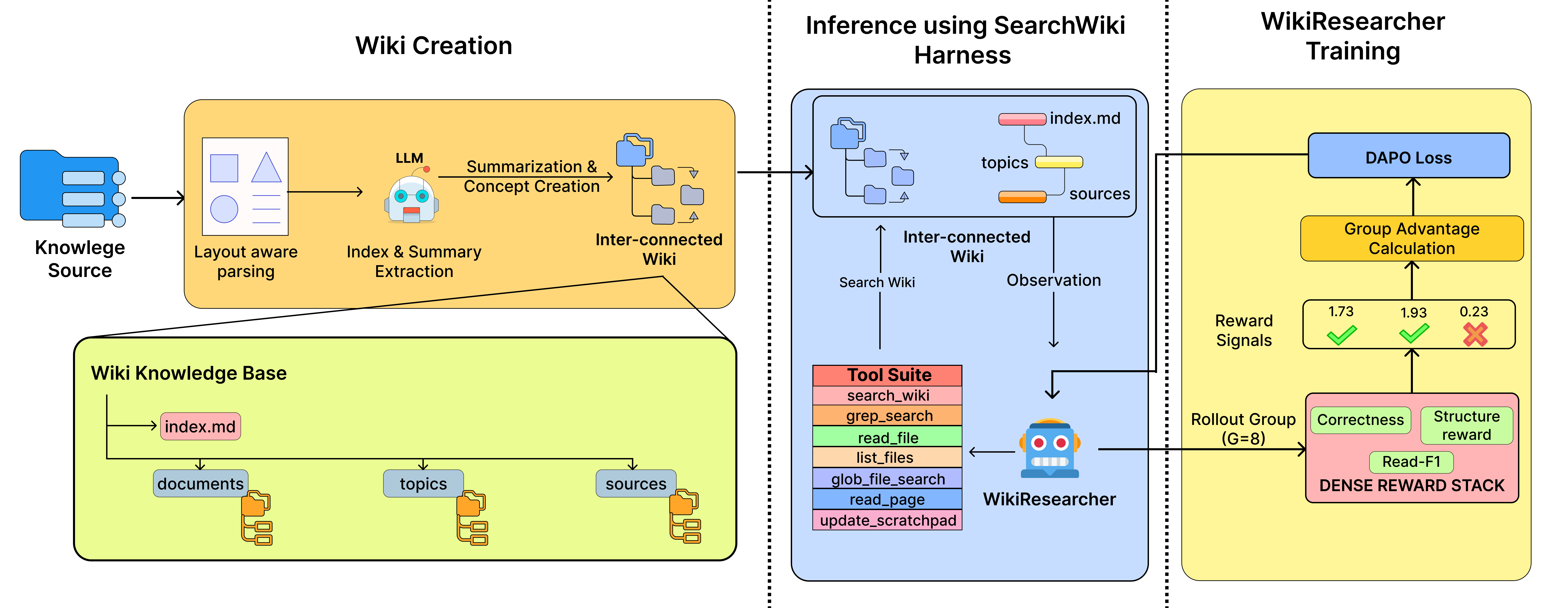}
\caption{\sys{} end-to-end framework.
Left: layout-aware parsing, TOC extraction, and joint summarization compile an unstructured knowledge source into a typed, three-layer wiki (\textit{documents/}, \textit{topics/}, \textit{sources/}).
Middle: given a query, the \sys{} harness navigates the wiki through a multi-turn tool-use loop (search, read, verify), returning a cite-tagged final answer.
Right: \ours{} is trained on rollout groups scored by the dense reward stack and optimized via DAPO.}
\label{fig:overview}
\end{figure*}

\sys{} is a two-stage framework for active information seeking over complex document corpora: (1) \textbf{Wiki Compilation}, which transforms an unstructured corpus into a typed, navigable three-layer wiki; and (2) \textbf{Corpus Navigation}, where an RL-tuned agent actively navigates the wiki via multi-turn tool interaction. Figure~\ref{fig:overview} depicts this as three stages - wiki creation, harness-based inference, and the offline RL training loop (Section~\ref{sec:training}) that produces the navigation policy used at inference.

Given a user query $q$ and a compiled \sys{} knowledge base $\mathcal{K}$, the agent interacts with the environment over $T$ steps to construct a decision trajectory:
\begin{equation}
\tau = (a_1, o_1, a_2, o_2, \ldots, a_T, o_T, y),
\end{equation}
where $a_t$ represents an assistant action (comprising chain-of-thought planning, scratchpad updates, or tool calls), $o_t$ is the environmental observation returned by the tool execution engine and appended to the context history~\cite{yao2023react}, and $y$ is the terminal answer produced when search terminates. Rather than relying on single-shot passage retrieval, $\pi_\theta$ dynamically decides which nodes to inspect, when to descend into page-level details, and when to backtrack based on intermediate observations.


\subsection{Wiki Compilation and Ingestion}
\sys{} wikis are built offline with a fully autonomous LLM compilation process.
Figure~\ref{fig:overview} (left) summarizes the path from a raw knowledge source to the typed wiki layout.

\subsubsection{Knowledge Representation.}
To preserve document hierarchy and cross-document relationships, \sys{} organizes a corpus $\mathcal{K}$ into a local wiki containing four node types::
\begin{enumerate}
    \item \textbf{Document overviews (\textit{documents/}):} one front-matter-typed summary per document (e.g., a Form~10-K overview) that lists section structure with page ranges and links into section sub-folders such as \textit{documents/doc\_id/\allowbreak part\_i/\allowbreak item\_1\_business.md}.
    \item \textbf{Topics (\textit{topics/}):} Cross-document concept pages (e.g., risk management policies) containing markdown links to related document overviews and cross-referenced topics.
    \item \textbf{Source pages (\textit{sources/}):} per-document JSON lists of page records (OCR/layout text, page index, and related metadata) loaded on demand via \textit{read\_page}.
    \item \textbf{Catalog files:} root and per-directory \textit{index.md} listings that map the tree, plus a build \textit{log.md}.
\end{enumerate}
Details about the directory layout has been added to the supplementary . 

\par The agent is not given the full corpus at once.
It discloses content through: search and catalog reads, inspect document or topic summaries, then call \textit{read\_page} for exact evidence.

This same wiki structure naturally extends to dialogue corpora by mapping session overviews to \textit{documents/} and turn records to \textit{sources/}. We train \ours{} on document-grounded ViDoRe wikis and demonstrate out-of-distribution generalization to dialogue-grounded memory benchmarks.

\subsubsection{Compilation Pipeline.}
The offline wiki construction process executes in three automated stages:

\paragraph{Stage 1: Layout-Aware Parsing.}
Document page images are processed with a vision-language OCR/layout pipeline to extract text and structure.
Short text inputs skip heavy visual parsing.
Parsed pages are serialized as page JSON records under \textit{sources/}, enabling downstream tools to target exact page ranges without re-parsing raw files. Short text inputs skip visual parsing.

\paragraph{Stage 2: TOC Extraction and Progressive Structuring.}
For long documents, our hierarchical indexer extracts a table of contents (TOC) that records section titles, nesting depth, and page spans.
That TOC is the blueprint for the on-disk tree under \textit{documents/}: the document root summary links to top-level sections; each section becomes a subdirectory with its own \textit{index.md} and section summary; nested TOC entries spawn further subdirectories and subsection summaries (sections within sections), each tied to the corresponding source page range.
Short documents without a usable TOC take a direct single-summary route.
Conversational memory sources follow the same principle at a different granularity: each conversation is compiled session by session, with every session becoming a short document and each turn a linked subsection tied back to the raw dialogue under \textit{sources/}.
The resulting hierarchy acts as a condensed information proxy: the agent can traverse Part~$\rightarrow$Item$\rightarrow$subsection summaries before calling \textit{read\_page} on the raw sources.

\paragraph{Stage 3: Joint Wiki Compiler.}
For each document unit, a single LLM call simultaneously generates (i)~TOC-aligned document and section summaries and (ii)~concept registration actions against a global topic registry. Performing summary generation and concept planning jointly minimizes information loss common in sequential pipelines and prevents duplicate topic creation. Compilation across independent documents is parallelized, while chronological sources are processed sequentially to maintain a shared topic registry.
Prompt constraints encourage canonical topic names and discourage near-duplicate concepts.

\begin{table}[htb!]
\centering
\tablesize\tablepad
\begin{tabular}{@{}lp{0.62\columnwidth}@{}}
\toprule
\textbf{Tool} & \textbf{Purpose} \\
\midrule
\textit{search\_wiki} & Pipe-split multi-query BM25 with RRF; returns top documents and page hints \\
\textit{grep\_search} & Regex scan over documents, topics, and sources \\
\textit{read\_file} & Read markdown summaries, sections, or topic pages \\
\textit{read\_page} & Read page-level source JSON for selected pages \\
\textit{list\_files} & List files in a directory \\
\textit{glob\_file\_search} & Match filenames by pattern \\
\textit{update\_scratchpad} & Search planning: persistent notes on sub-questions and progress \\
\bottomrule
\end{tabular}
\caption{\sys{} tool suite and purpose used in training and evaluation.}
\label{tab:tools}
\end{table}

\subsection{Wiki Navigation and Policy Learning}
\subsubsection{Tool-Based Navigation.}

We list seven tools available to the agent in Table~\ref{tab:tools}.
The system prompt requires \textit{update\_scratchpad} on every tool turn, in parallel with other calls for persistent planning.



\paragraph{Output protocol.}
Each assistant turn follows a Qwen3.5-style XML protocol~\cite{team2026qwen35}: an optional \textit{<think>} block for reasoning, zero or more \textit{<tool\_call>} blocks using the native \textit{<function=...>} parameter format, or a terminal \textit{<answer>} block. Final answers must conclude with explicit citations in the form \textit{[Relevant Documents: DOC\_NAME[pages]]}. Trajectory format compliance is validated via automated tag checking, with malformed outputs down-weighted during training. Full multi-turn trajectory examples are included in the supplementary material.

\subsubsection{Policy Optimization via Reinforcement Learning.}
\label{sec:training}

We train the navigation policy $\pi_\theta$ online: for each training query, the agent generates multi-turn tool-use trajectories against the corresponding wiki environment. Our RL alignment strategy rests on the following core components:

\paragraph{Online rollouts.}
Each step samples training queries and generates $G=8$ completions per prompt.
A multi-turn scheduler executes tool calls against the split-local wiki and appends observations to the context dialogue.

\paragraph{Retrieved Token Masking.}
Tool responses $o_t$ are included in the context history but excluded from policy loss calculation~\cite{jin2025searchr1}. Gradient updates are restricted exclusively to model-generated tokens (planning, tool calls, and answers), preventing the policy from imitating retrieved OCR text.

\paragraph{DAPO objective.}
For each query we sample a group of $G$ trajectories with scalar rewards $\{R_i\}_{i=1}^{G}$ and form group-relative advantages
\begin{equation}
\hat{A}_{i}=\frac{R_i-\mathrm{mean}(\{R_j\}_{j=1}^{G})}{\mathrm{std}(\{R_j\}_{j=1}^{G})+\delta},
\qquad \delta=10^{-8}.
\end{equation}
Following DAPO~\cite{yu2025dapo}, we maximize the token-level clipped surrogate with asymmetric clip bounds
\begin{multline}
J_{\mathrm{DAPO}}(\theta)
=
\mathbb{E}\!\left[
\frac{1}{Z}
\sum_{i=1}^{G}\sum_{t=1}^{|o_i|}
\min\!\bigl(
r_{i,t}(\theta)\,\hat{A}_{i},\right.\\
\left.\mathrm{clip}\bigl(r_{i,t}(\theta),\,1{-}\varepsilon_{\mathrm{low}},\,1{+}\varepsilon_{\mathrm{high}}\bigr)\,\hat{A}_{i}
\bigr)
\right],
\end{multline}
where $Z=\sum_{i=1}^{G}|o_i|$, the expectation is over $(q,\{o_i\})\sim\mathcal{D}$, and the importance ratio is
\begin{equation}
r_{i,t}(\theta)=\frac{\pi_{\theta}(o_{i,t}\mid q,o_{i,<t})}{\pi_{\theta_{\mathrm{old}}}(o_{i,t}\mid q,o_{i,<t})}.
\end{equation}
We set $\varepsilon_{\mathrm{low}}=0.2$ and $\varepsilon_{\mathrm{high}}=0.28$ (Clip-Higher).
Because Retrieved Token Masking zeros the contribution of tool-observation tokens, the inner sum runs only over model-generated tokens (planning, tool calls, and answers).
We use DAPO for the policy update and GDPO~\cite{liu2026gdpo} for group reward scaling under the same reward stack.

\paragraph{Two-stage training.}
Training proceeds in two phases. In Stage~1, the policy is optimized end-to-end using the full search and QA reward stack. In Stage~2, we restrict gradient updates strictly to final answer tokens while retaining the QA correctness reward. Training with QA rewards alone without this token mask collapses navigation policy learning; this two-stage, answer-masked recipe prevents trajectory degradation over long search chains.

\paragraph{Dense reward stack.}
The scalar reward combines answer quality, retrieval quality, and trajectory discipline in a weighted sum formulation:
\begin{itemize}
    \item \textbf{Correctness}: LLM as a Judge for scoring correctness - $1.0$ for Correct, $0.3$ for partial and $0$ for incorrect ~\cite{zheng2023llmjudge,liu2023geval}.
    \item \textbf{Read-F1} and \textbf{retrieval rank / NDCG}: reward reading and ranking gold pages/documents by citation and reads~\cite{jarvelin2002ndcg}.
    \item \textbf{Structural Rewards:} format compliance, discovery-phase traversal heuristics, search-tool usage, and parallel-tool bonuses; plus penalties for excess turns, duplicate calls, answering without tools, or non-submitting finals.
\end{itemize}
Exact component weights have been listed in the supplementary material.
The reward mixture was calibrated through analysis of failure modes: correctness alone risks reward guessing, retrieval alone incentivizes reading without answering, and structural shaping alone encourages busywork~\cite{ng1999rewardshaping,amodei2016concrete,skalse2022rewardhacking}. Rollouts truncated by context limits or turn bounds receive a hard total reward of zero ($R=0$), enforcing budget discipline during exploration.

\section{Experimental Setup}
\label{sec:experiments}

\subsection{Training Setup}

\paragraph{Synthetic Data Generation.}

To train our RL policy, we utilize a three-stage synthetic data generation pipeline inspired by ViDoRe.
First, for each gold evidence unit, we prompt a teacher LLM to generate candidate questions grounded in the underlying selected source text.
After generation, candidates are passed through a deterministic filter, which includes regular expressions, length validation, coverage verification and checks to eliminate trivial queries containing direct answer leakage.
Following this, we evaluate the remaining queries by an LLM judge across four critical rubrics: factual faithfulness, query clarity, absence of answer leakage, and adherence to formatting style. A candidate is discarded if it fails to follow any criterion of the four. This filtering resulted in retaining approximately $25\%$ of the total generated queries, yielding a final dataset of $1{,}800$ high quality instances. Each domain split contains $225$ queries, and we divide our data into $1{,}600$ training samples and $200$ validation samples.



\subsubsection{Implementation Details.}

We initialize \ours{} from Qwen3.5-9B and train with DeepSpeed ZeRO-3~\cite{rajbhandari2020zero}, sequence parallelism of $4$ across $8$ A100 GPUs, and a colocated vLLM engine for generation, optimizing over $1{,}600$ synthetic queries for one epoch (${\approx}400$ steps) at a peak learning rate of $10^{-6}$. Full training configuration is in the supplementary material.

\subsubsection{Context Budget and Reasoning Length.}

Long CoT reasoning inflates context usage across multi-turn trajectories, pushing rollouts toward the $65{,}536$-token limit; since truncated trajectories receive zero reward, verbose thinking can destabilize training despite appearing high-quality. We therefore favor concise step-by-step planning paired with the scratchpad tool over unconstrained reasoning traces.

\subsection{Evaluation Framework}

We evaluate performance along two primary axes: answer quality, assessed via an LLM-as-judge protocol, and retrieval quality, measured through page- and document-level ranking metrics. Crucially, all evaluations execute under the identical agent scheduler, tool suite, and environment configuration used during training. This guarantees that observed gains reflect genuine improvements in the learned navigation policy rather than artifacts of a mismatched test harness.

\paragraph{Datasets and Benchmarks.}
We evaluate in-distribution performance on all eight \emph{ViDoRe~v3} domains. For out-of-distribution transfer, we evaluate on \emph{FinanceBench} (financial QA) and three dialogue memory benchmarks: \emph{LoCoMo}, \emph{LongMemEval}, and \emph{PersonaMem-v2}~\cite{vidorev3,islam2023financebench,maharana2024locomo,wu2024longmemeval,jiang2025personamemv2}. While evaluating on memory benchmarks, we adopt NapMem's exact setup (user-disjoint $60/20/20$ splits, token F1, LLM-judge, and multiple-choice accuracy) for direct comparability~\cite{xu2026napmem}.

\paragraph{Answer Correctness Metrics.}
Answer quality is evaluated using GPT-OSS-120B~\cite{openai2025gptoss} as an external LLM judge. For \emph{ViDoRe~v3}, we adopt the official ViDoRe judge prompt~\cite{vidorev3}, which categorizes answers as Correct, Partially Correct, or Incorrect. For out-of-distribution holdouts, we use NapMem's binary Correct/Incorrect judge prompt (or multiple-choice accuracy where applicable). Across all open-ended tasks, we report strict accuracy—counting only fully Correct answers—both per split and macro-averaged across domains. Full judge templates and prompts are included in the supplementary material.

\paragraph{Retrieval and Ranking Metrics.} 
Retrieval quality uses Recall@5 and nDCG@5 at page and document levels. Rankings follow a cite-then-read convention: cited evidence ranks first, followed by pages/documents accessed via read\_page or read\_file. Page-level metrics may understate performance when agents answer from summaries or topic pages without accessing the exact source; document-level metrics are more robust to this.

\paragraph{Comparative Baselines.}
All evaluated models operate under the identical \sys{} tool suite and environment harness. We compare our RL-tuned policy, \ours{}, against its same-size untrained base model (Qwen3.5-9B) as well as several larger baseline models: Qwen3.6-27B, Qwen3.5-397B, Gemma-4-31B, and MiniMax-M2.5~\cite{team2026qwen35,gemma2026gemma4,minimax2026m25}.

\begin{table*}[htb]
\centering
\tablesize\tablepad
\begin{tabular}{@{}lccccccccc@{}}
\toprule
& \multicolumn{5}{c}{English Datasets} & \multicolumn{3}{c}{French Datasets} & \\
\cmidrule(lr){2-6}\cmidrule(lr){7-9}
\textbf{Model} & \textbf{C.S.} & \textbf{Fin.} & \textbf{Phar.} & \textbf{H.R.} & \textbf{Ind.} & \textbf{Phys.} & \textbf{Ener.} & \textbf{Fin.} & \textbf{Avg.} \\
\midrule
Gemma-4-31B$^\dagger$
  & 81.47 & 55.45 & 60.83 & 56.39 & 52.36
  & 74.94 & 58.82 & 51.46 & $61.47{\pm}0.79$ \\
MiniMax-M2.5$^\dagger$
  & 71.63 & 20.23 & 35.16 & 24.11 & 23.97
  & 70.42 & 29.33 & 23.28 & $37.27{\pm}0.72$ \\
Qwen3.5-397B$^\dagger$
  & 86.67 & 63.54 & 72.39 & 64.41 & 58.78
  & 83.39 & 69.16 & 57.76 & $69.51{\pm}0.73$ \\
Qwen3.6-27B$^\dagger$
  & \underline{87.36} & 65.43 & 71.52 & \underline{68.13} & \underline{61.84}
  & 82.67 & 71.05 & \underline{59.48} & $70.94{\pm}0.72$ \\
Qwen3.5-9B$^\dagger$
  & 82.33 & 57.61 & 67.81 & 54.72 & 51.83
  & 79.69 & 63.31 & 54.32 & $63.95{\pm}0.76$ \\
\textbf{\ours}
  & \textbf{86.51} & \textbf{\underline{65.80}} & \textbf{\underline{77.84}} & \textbf{63.94} & \textbf{61.01}
  & \textbf{\underline{83.94}} & \textbf{\underline{74.03}} & \textbf{57.71} & $\textbf{\underline{71.35}}{\pm}0.73$ \\
\bottomrule
\end{tabular}
\caption{ViDoRe~v3 strict QA accuracy (\%; same \sys{} harness).
Avg.\ is the unweighted macro average over eight splits; ${\pm}$ is the 95\% CI half-width ($n_{\mathrm{boot}}{=}10^{4}$; per-split results in supplementary); $\dagger$=untrained.
\ours{} is bold; best per column is underlined.}
\label{tab:vidore-qa}
\end{table*}

\begin{table*}[htb]
\centering
\small
\tablesize\tablepad
\begin{tabular}{@{}lccccc@{}}
\toprule
\textbf{Method} & \multicolumn{2}{c}{\textbf{LoCoMo}} & \multicolumn{2}{c}{\textbf{LongMemEval}} & \textbf{PersonaMem-v2} \\
\cmidrule(lr){2-3}\cmidrule(lr){4-5}\cmidrule(lr){6-6}
& F1 & LLM-J & F1 & LLM-J & Acc. \\
\midrule
Zep$^\ddagger$ & $36.09{\pm}0.38$ & $52.21{\pm}0.31$ & $52.61{\pm}0.20$ & $73.33{\pm}0.94$ & $36.33{\pm}0.81$ \\
MemoryOS$^\ddagger$ & $31.40{\pm}0.53$ & $44.28{\pm}1.22$ & $22.68{\pm}0.30$ & $23.67{\pm}0.58$ & $35.82{\pm}1.87$ \\
MemOS$^\ddagger$ & $35.22{\pm}0.38$ & $55.82{\pm}0.65$ & $39.16{\pm}0.40$ & $54.33{\pm}0.47$ & $37.58{\pm}0.44$ \\
AgeMem$^\ddagger$ & $38.00{\pm}0.12$ & $45.02{\pm}0.12$ & $37.83{\pm}1.38$ & $51.33{\pm}0.94$ & $24.19{\pm}0.67$ \\
Mem0$^\ddagger$ & $41.19{\pm}0.36$ & $60.86{\pm}0.38$ & $53.86{\pm}0.54$ & $78.00{\pm}0.82$ & $38.89{\pm}0.34$ \\
NapMem-397B (Untrained)$^\ddagger$ & $38.31{\pm}0.13$ & $54.42{\pm}0.81$ & $53.85{\pm}1.25$ & $78.33{\pm}2.08$ & $46.10{\pm}1.15$ \\
NapMem-9B w/ RL$^\ddagger$ & $41.28{\pm}0.59$ & $59.92{\pm}0.78$ & $57.41{\pm}1.12$ & $80.33{\pm}0.47$ & $47.97{\pm}1.01$ \\
\midrule
Gemma-4-31B$^\dagger$ & $48.04{\pm}0.50$ & $58.99{\pm}0.13$ & $54.45{\pm}0.15$ & $67.50{\pm}2.50$ & $46.87{\pm}0.66$ \\
MiniMax-M2.5$^\dagger$ & $24.60{\pm}1.23$ & $33.28{\pm}1.51$ & $22.41{\pm}2.76$ & $30.50{\pm}2.50$ & $34.36{\pm}0.33$ \\
Qwen3.5-397B$^\dagger$ & $52.10{\pm}0.16$ & $63.87{\pm}0.23$ & $56.22{\pm}0.80$ & $69.50{\pm}0.50$ & $50.00{\pm}0.82$ \\
Qwen3.6-27B$^\dagger$ & \underline{$55.84{\pm}1.65$} & \underline{$69.03{\pm}1.89$} & \underline{$61.70{\pm}2.58$} & $77.50{\pm}3.50$ & \underline{$57.08{\pm}3.18$} \\
\textbf{\ours} & {\boldmath $48.31{\pm}0.18$} & {\boldmath $63.65{\pm}0.10$} & {\boldmath $59.81{\pm}1.46$} & {\boldmath \underline{$80.67{\pm}2.50$}} & {\boldmath $47.26{\pm}0.05$} \\
\bottomrule
\end{tabular}
\caption{Memory-intensive task performance on LoCoMo, LongMemEval, and PersonaMem-v2 (\%; mean${\pm}$uncertainty).
($^\ddagger$=results from NapMem~\cite{xu2026napmem}); ($\dagger$=untrained).
Uncertainty details in the supplementary.
\ours{} is bold; best per column is underlined.}
\label{tab:memory}
\end{table*}
\section{Results and Analysis}
\label{sec:result_analysis}
\subsection{Results on ViDoRe v3}

\textbf{\ours{} outperforms other baselines on end-to-end QA accuracy:}
We report the per-split QA accuracy under the official ViDoRe LLM-as-judge prompt~\cite{vidorev3} (judge model: GPT-OSS-120B) in Table~\ref{tab:vidore-qa}.
\ours{} leads at $71.35{\pm}0.73$, ahead of untrained Qwen3.6-27B ($70.94{\pm}0.72$) and Qwen3.5-397B ($69.51{\pm}0.73$), and improves $+7.4$ points over the same-size Qwen3.5-9B ($63.95{\pm}0.76$).
Paired bootstrap comparisons put the 9B and 397B gains well clear of zero; the edge over 27B is smaller but still favors \ours{} (details in Supplementary material).

\textbf{\ours{} leads retrieval metrics across domains.}
We summarize the cite-then-read Recall@$5$ and nDCG@$5$ at page and document level (macro average over splits) in Table~\ref{tab:vidore-ret-summary} .
\ours{} improves page-level nDCG@$5$ over both untrained models, consistent with our retrieval-aligned reward stack, and remains competitive at the document level -- notably surpassing Qwen3.5-397B, a same-family model with $44\times$ the parameters, on page-level recall and nDCG. Per-domain breakdowns are in the supplementary material.

\begin{table}[htb]
\centering
\tablesize\tablepad
\begin{tabular}{@{}lcccc@{}}
\toprule
& \multicolumn{2}{c}{Page} & \multicolumn{2}{c}{Doc} \\
\cmidrule(lr){2-3}\cmidrule(lr){4-5}
\textbf{Model} & \textbf{nDCG@5} & \textbf{R@5} & \textbf{nDCG@5} & \textbf{R@5} \\
\midrule
Gemma-4-31B$^\dagger$ & 43.82 & 38.16 & 82.89 & 85.02 \\
MiniMax-M2.5$^\dagger$ & 20.84 & 18.35 & 48.14 & 50.09 \\
Qwen3.5-397B$^\dagger$ & 47.15 & 43.21 & \underline{87.66} & \underline{91.33} \\
Qwen3.6-27B$^\dagger$ & 47.05 & 43.70 & 87.41 & 90.41 \\
Qwen3.5-9B$^\dagger$ & 40.55 & 37.16 & 84.23 & 89.35 \\
\textbf{\ours} & \textbf{\underline{53.22}} & \textbf{\underline{48.55}} & \textbf{86.83} & \textbf{90.02} \\
\bottomrule
\end{tabular}
\caption{ViDoRe~v3 cite-then-read retrieval (\%), averaged over splits.
Same models/order/harness as Table~\ref{tab:vidore-qa}; $\dagger$=untrained.
\ours{} is bold; best per column is underlined.}
\label{tab:vidore-ret-summary}
\end{table}

\subsection{Performance on Memory Benchmarks}
We report LoCoMo, LongMemEval, and PersonaMem-v2 under NapMem's evaluation protocol in Table \ref{tab:memory} (with same 60/20/20 user-level splits, judge prompt, and F1 / LLM-J metrics), so gaps vs.\ NapMem are not an artifact of mismatched splits or rubrics.
NapMem's strongest reported 9B agent is trained specifically for user-memory navigation on these suites~\cite{xu2026napmem}.
\sys{} is not: harness rows reuse the same wiki tools as ViDoRe, and \ours{} is the ViDoRe-trained 9B student.
Still, \sys{} leads NapMem on LoCoMo F1 and LLM-judge (27B harness), and \ours{} is competitive on LongMemEval LLM-J (80.67 vs.\ NapMem's 80.33) despite no memory-specific training.
The untrained 27B harness also exceeds NapMem on PersonaMem-v2, indicating that a general wiki navigator can match or beat a memory-specialized policy without memory-specific training.

\subsection{Generalization to FinanceBench}
We report the results on FinanceBench dataset in Table~\ref{tab:financebench}. \ours{} reaches 83.33\% accuracy, well above the leaderboard baselines from \citet{islam2023financebench} (GPT-4-Turbo, GPT-4, Llama-2, all $\leq$19.30\%) and close to the GPT-4-Turbo \textit{Oracle} upper bound (85.33\%), despite using retrieved rather than gold evidence. Within our own harness, where all models share identical retrieval, \ours{} trails the best system, Qwen3.6-27B$^\dagger$ (85.33\%), by only 2 points, and Qwen3.5-397B (84.67\%) by 1.34 -- notable given \ours{} is trained solely on ViDoRe. This out-of-distribution transfer suggests its evidence-localization and wiki-traversal skills generalize well beyond the training domain.

\begin{table}[htb]
\centering
\tablesize\tablepad
\begin{tabular}{@{}lc@{}}
\toprule
\textbf{Model / Setting (\textit{Shared Store})} & \textbf{Acc.\ (\%)} \\
\midrule
GPT-4-Turbo(\textit{Oracle})$^\ddagger$ & 85.33 \\
GPT-4-Turbo$^\ddagger$ & 19.30 \\
Llama-2$^\ddagger$ & 19.30 \\
GPT-4$^\ddagger$ & 16.70 \\
\midrule
Gemma-4-31B$^\dagger$ & 78.00 \\
MiniMax-M2.5$^\dagger$ & 29.33 \\
Qwen3.5-397B$^\dagger$ & 84.67 \\
Qwen3.5-9B$^\dagger$ & 79.33 \\
Qwen3.6-27B$^\dagger$ & \underline{85.33} \\
\textbf{\ours}   & \textbf{83.33} \\
\bottomrule
\end{tabular}
\caption{FinanceBench Shared-Store results (\%).
($^\ddagger$ numbers from~\cite{islam2023financebench}.)
Bottom block: same \sys{} harness and model order as Table~\ref{tab:vidore-qa}.$\dagger$=untrained.
\ours{} is bold; best in the harness block is underlined.}
\label{tab:financebench}
\end{table}

\subsection{Ablations}
We disentangle the contributions of our hierarchical wiki representation and multi-turn tool use in Table~\ref{tab:vidore-ablate}, while comparing against passive single-turn RAG baselines.

We consider two ablations of \sys{}. (1) \textit{w/o typed hierarchy} retains the full tool suite but exposes the agent only to a flat collection of source pages, removing the topic and document layers from the wiki. (2) \textit{w/o tools (BM25)} removes multi-turn interaction entirely, instead retrieving BM25 top-$k$ OCR chunks and generating an answer in a single pass. We additionally report a stronger passive baseline, \textit{dense RAG}, which retrieves the top-10 pages using \textit{nvidia/nemotron-colembed-vl-8b-v2} and answers in one turn with the same untrained 9B/27B generators.

The results show consistent gains from both hierarchy and interaction. With the untrained 27B model, full \sys{} achieves 70.94, outperforming the flat-hierarchy agent (65.47), dense RAG (62.61), and BM25 RAG (29.60). The same trend holds with the untrained 9B model, where \sys{} reaches 63.95, compared to 57.30 for dense RAG and 50.39 without the typed hierarchy. Applying \ours{} further improves the 9B model to 71.35, demonstrating that reinforcement learning provides gains beyond model scaling, while the hierarchical representation and multi-turn interaction each contribute independently beyond simply replacing BM25 with a stronger dense retriever.

These improvements stem from the different search paradigm enabled by \sys{}. In passive RAG, recovery from an initial retrieval miss is largely restricted to reformulating retrieval queries while continuing to interact blindly with the corpus. In contrast, \sys{} treats recovery as an explicit decision process: the agent can iteratively revise queries, inspect summaries, traverse topic links, search for literals with \textit{grep}, and finally verify evidence using \textit{read\_page}. The typed hierarchy is central to this process, as topic and summary pages provide inexpensive high-level context that guides exploration before the agent commits to reading source pages. Rather than serving solely as an improved retrieval index, the wiki becomes a navigable action space that supports efficient information seeking.

\begin{table}[t]
\centering
\tablesize\tablepad
\begin{tabular}{@{}lr@{}}
\toprule
\textbf{Setting} & \textbf{Avg.\ QA} \\
\midrule
\textbf{Full \sys{} (27B)} & 70.94 \\
\quad\textit{w/o typed hierarchy} & 65.47 \\
\quad\textit{dense RAG (Dense, single-turn)} & 62.61 \\
\quad\textit{w/o tools (BM25, single-turn)} & 29.60 \\
\midrule
\textbf{Full \sys{} (9B, w/o RL)} & 63.95 \\
\quad\textit{w/o typed hierarchy} & 50.39 \\
\quad\textit{dense RAG (Dense, single-turn)} & 57.30 \\
\midrule
\textbf{\ours} & \textbf{\underline{71.35}} \\
\bottomrule
\end{tabular}
\caption{ViDoRe~v3 ablations (macro-average QA \%; judge: GPT-OSS-120B).
Dense RAG uses \textit{nvidia/nemotron-colembed-vl-8b-v2} (top-$10$).
$\dagger$=untrained.
\ours{} is bold; best score is underlined.}
\label{tab:vidore-ablate}
\end{table}
\section{Conclusion}
\label{sec:conclusion}
In this paper, we introduced \sys, a two-stage framework that replaces passive flat RAG with active corpus navigation. First, \sys{} automatically compiles unstructured corpora into a typed, three-layer wiki comprising document overviews, cross-document topic pages, and page-level source records. Second, we train \ours{} via online RL to master multi-turn tool use, evidence verification, and dynamic backtracking. Our empirical results strongly validate this paradigm: \ours{} ranks first overall on ViDoRe~v3 in end-to-end QA accuracy ($71.35\%$)—outperforming models up to $44\times$ its parameter scale - while dominating page-level retrieval ranking ($53.22\%$ nDCG@5). On FinanceBench, \ours{} achieves $83.33\%$ accuracy - approaching the Oracle upper bound ($85.33\%$). Finally, it surpasses domain-specialized memory policies on LoCoMo, PersonaMem-v2 and LongMemEval without any memory-specific training, establishing learned wiki navigation as a superior and generalizable alternative to chunk-based retrieval.



\begin{small}
\bibliography{references}
\end{small}

\end{document}